\documentclass[lettersize,journal]{IEEEtran}
\usepackage{amsmath,amsfonts}
\usepackage{algorithmic}
\usepackage{algorithm}
\usepackage{array}
\usepackage{booktabs}
\usepackage[caption=false,font=normalsize,labelfont=sf,textfont=sf]{subfig}
\usepackage{textcomp}
\usepackage{stfloats}
\usepackage{multirow}
\usepackage{url}
\usepackage{verbatim}
\usepackage{graphicx}
\usepackage{amssymb}
\usepackage{booktabs}
\usepackage{amssymb}
\usepackage{bm}
\usepackage[colorlinks,linkcolor=blue]{hyperref}

\usepackage{cite}
\begin{document}

\title{A Transformer-Based Feature Segmentation and Region Alignment Method For UAV-View Geo-Localization}

\author{
Ming Dai, Jianhong Hu, Jiedong Zhuang, Enhui Zheng
\thanks{Ming Dai, Jianhong Hu, Jiedong Zhuang, Enhui Zheng are with the Unmanned System Application Technology Research Institute, China Jiliang University, Hangzhou 310018, China (email: s20010802003@cjlu.edu.cn; zjuhjh@126.com; p1901085206@cjlu.edu.cn; ehzheng@cjlu.edu.cn). Enhui Zheng is the Corresponding Author.}
}

\IEEEpubid{\begin{minipage}{\textwidth}\ \\[12pt] \centering
		\ \\
		\ \\
		\ \\
		Copyright © 2021 IEEE. Personal use of this material is permitted. However, permission to use this material for any other\\
		purposes must be obtained from the IEEE by sending an email to pubs-permissions@ieee.org.
\end{minipage}}

\maketitle

\begin{abstract}
Cross-view geo-localization is a task of matching the same geographic image from different views, e.g., unmanned aerial vehicle (UAV) and satellite. The most difficult challenges are the position shift and the uncertainty of distance and scale. Existing methods are mainly aimed at digging for more comprehensive fine-grained information. However, it underestimates the importance of extracting robust feature representation and the impact of feature alignment. The CNN-based methods have achieved great success in cross-view geo-localization. However it still has some limitations, e.g., it can only extract part of the information in the neighborhood and some scale reduction operations will make some fine-grained information lost. In particular, we introduce a simple and efficient transformer-based structure called Feature Segmentation and Region Alignment (FSRA) to enhance the model's ability to understand contextual information as well as to understand the distribution of instances. Without using additional supervisory information, FSRA divides regions based on the heat distribution of the transformer's feature map, and then aligns multiple specific regions in different views one on one. Finally, FSRA integrates each region into a set of feature representations. The difference is that FSRA does not divide regions manually, but automatically based on the heat distribution of the feature map. So that specific instances can still be divided and aligned when there are significant shifts and scale changes in the image. In addition, a multiple sampling strategy is proposed to overcome the disparity in the number of satellite images and that of images from other sources. Experiments show that the proposed method has superior performance and achieves the state-of-the-art in both tasks of drone view target localization and drone navigation. Code will be released at \href{https://github.com/Dmmm1997/FSRA}{https://github.com/Dmmm1997/FSRA}
\end{abstract}


\section{Introduction}
\IEEEPARstart{C}{ROSS-VIEW} geo-localization aims to match an image from one perspective to the most similar image from another perspective that represents the same geographic target. Its essence can be understood as a retrieval task of images from two different sources. Cross-view geo-localization can be applied to many fields such as agriculture, aerial photography, autonomous vehicles, drone navigation, event detection, accurate delivery, and so on \cite{ref1},\cite{ref2},\cite{ref3},\cite{ref4},\cite{ref5}. The predecessors did a lot of arduous work \cite{ref6}, \cite{ref7}, \cite{ref8}, \cite{ref9}, mostly studying the matching of ground panoramic images and satellite images. However, the intervention of the drone-view will further expand the application of cross-view geo-localization \cite{ref10} \cite{ref51}. The application of matching UAVs and satellite images can be roughly divided into the following two types: \textbf{Drone view target localization and Drone navigation}. For example, the image acquired by UAVs is used to match the satellite image of the same geographic location. Generally, the satellite image contains precise GPS coordinate information. Indirectly, UAVs can be located in real-time by adopting the geographical information from matched satellite images and the navigation of the drone can be realized without GPS equipment.

\begin{figure}[!t]
\centering
\includegraphics[width=0.5\textwidth]{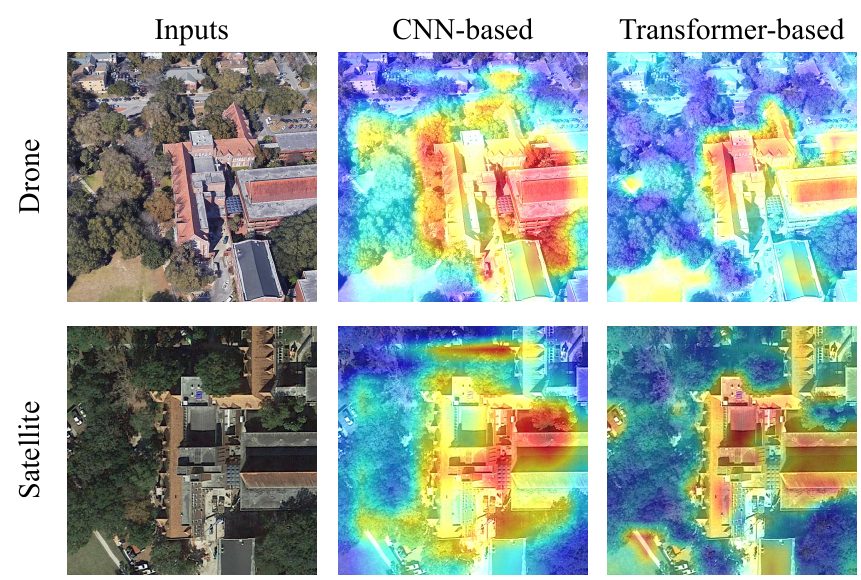}
\caption{The images on the left column is the input image from drone-view and satellite-view. The images in the middle column is the heatmap of CNN-based state-of-the-art network LPN \cite{ref11}. The images on the right column is the heatmap of our Transformer-based strong baseline.}
\label{fig_1}
\end{figure}

In recent years, Due to the rapid development of deep learning, significant progress has been made in cross-view geo-localization. By observing the CNN-based method, we found two potential problems. (I) Cross-view geo-localization needs to dig out the relevant information between contexts. Images from different domains have positional transformations such as rotation, scale, and offset. Therefore, fully understanding the semantic information of the global context is necessary. However, CNN-based methods mainly focus on small discriminative regions due to a Gaussian distribution of effective receptive fields \cite{ref12}. Given the limitations of the pure CNN-based methods \cite{ref13}, the attention modules have been introduced to explore long-range relationships\cite{ref14}. However, most of the methods embed the attention mechanism into the deep convolutional network, which enhances contextual connections to a certain extent. (II) Fine-grained information is very important for the task of retrieval. The down-sampling operations i.e., pooling and stride convolution of the CNN-based method can reduce the resolution of the image, while invisibly destroying the recognizable fine-grained information. In view of this, Transformer as a strong context-sensitive information extractor will have a role to play in Cross-View Geo-Localization.

\begin{figure*}[!t]
\centering
\includegraphics[width=1.0\textwidth]{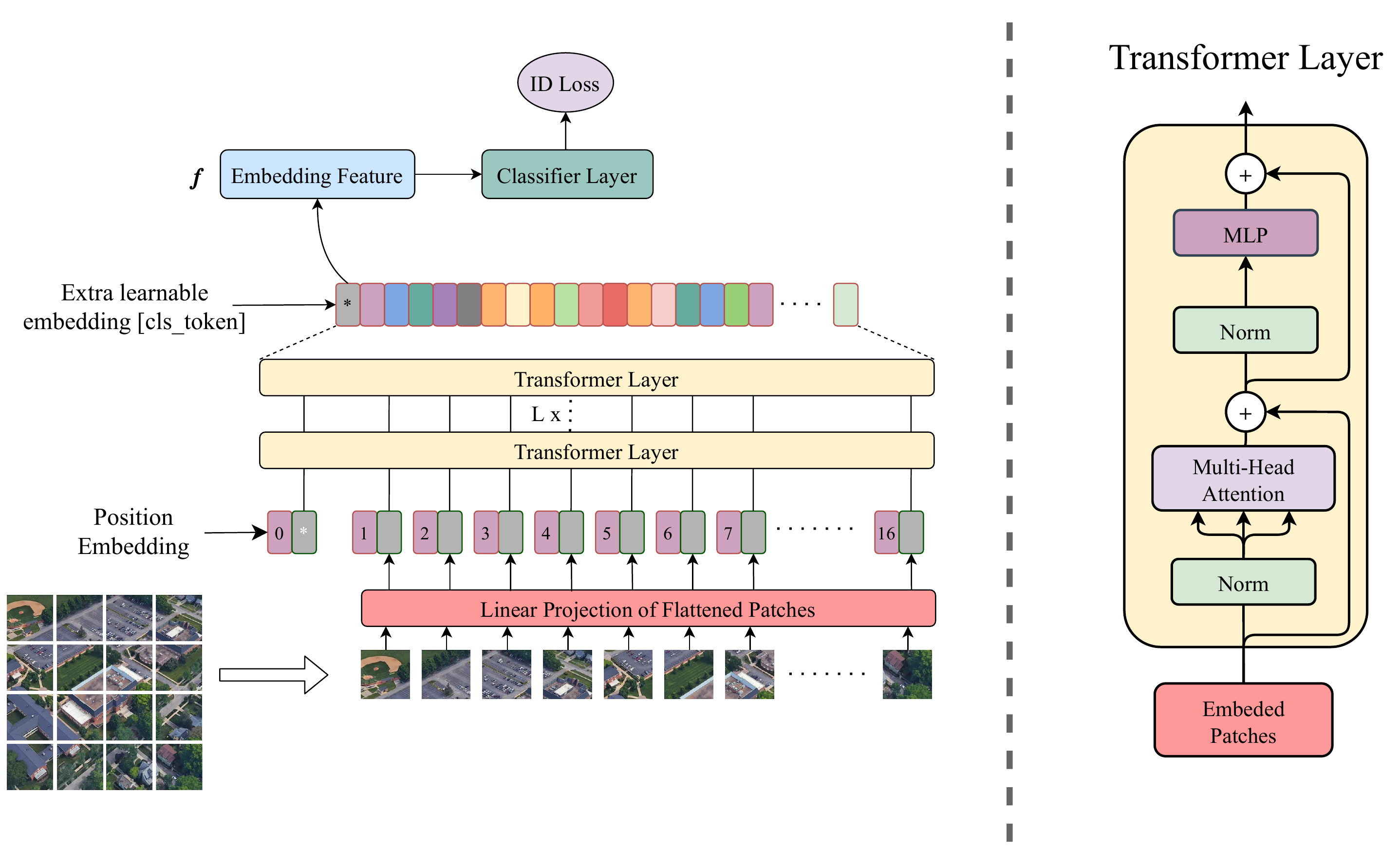}
\caption{Transformer-based strong baseline framework. Output $[cls\_token]$ marked with $*$ is served as the global feature $f$. $Classifier Layer$ contains linear layer, relu, batchnorm1d and dropout. ID Loss represents CrossEntropy loss without label-smooth. In addition, we provide a simplified Transformer Layer structure, the specific structure can be found in Vit \cite{ref15}.}
\label{fig_2}
\end{figure*}

In order to improve the visibility of model performance, we draw the heatmaps regarding Grad-CAM \cite{ref16}, as in Fig. \ref{fig_1}. The heatmaps come from the output of the last attention layer of the Vit but excluding the patch of learnable embedding. However, the output of the Transformer has only 3 dimensions, and we reduce the dimension of patches to the original image dimension by the inverse method of flattening. Thus the results of Vit concerns are visualized. We compare the heatmaps between the state-of-the-art CNN-based method LPN \cite{ref11} and our transformer-based strong baseline. Compared to the CNN-based method, the Transformer-based method can more clearly identify salient features such as buildings and roads and ignore background information such as trees.  

\begin{figure*}[!t]
\centering
\includegraphics[width=1.0\textwidth]{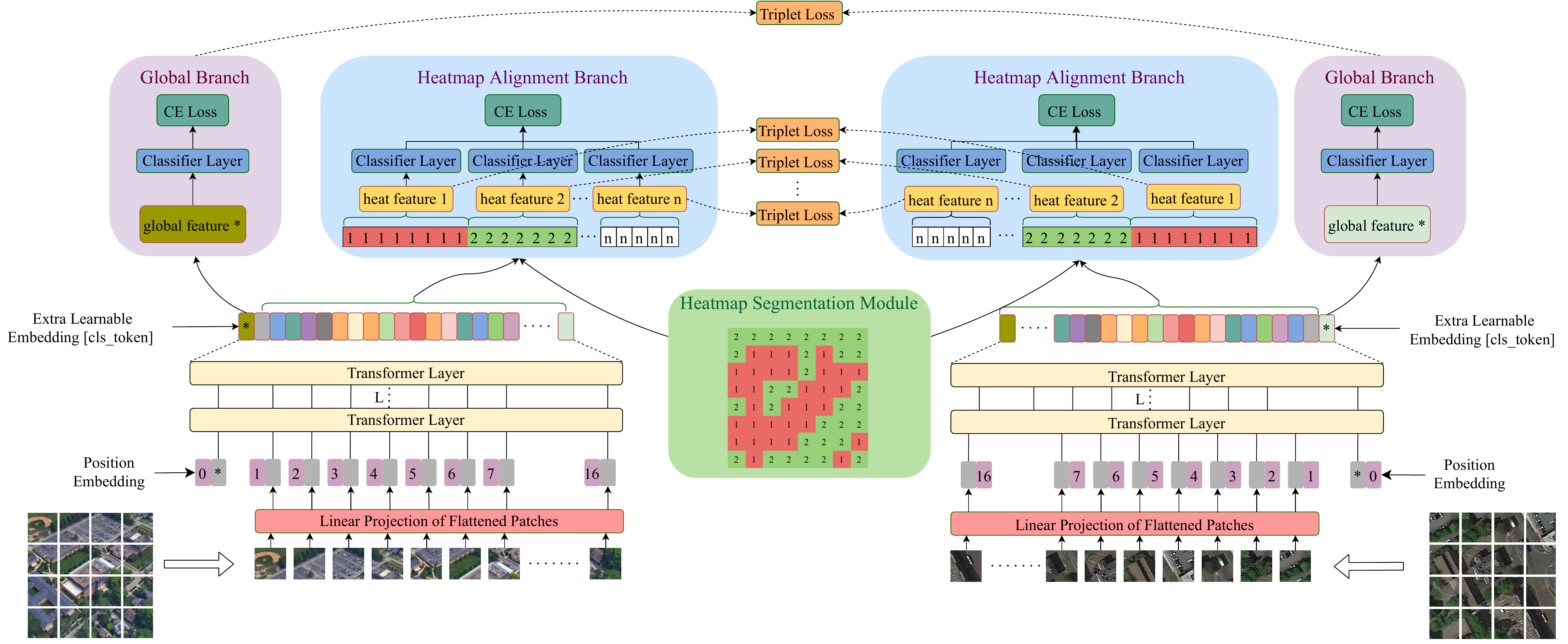}
\caption{The framework of proposed FSRA. The Heatmap Segmentation Module (light green) reorders the heatmap information and evenly divides it into regions according to the distribution of the heatmap to achieve the purpose of segmenting different characteristic content. Heatmap Alignment Branch (light blue) pools the features of each region to obtain feature vectors and performs classification supervision on each feature vector. In order to achieve end-to-end learning, TripletLoss is applied to each branch to narrow the distance of the same feature content. FSRA also retains the global branch (light purple) of the transformer-based strong baseline.}
\label{fig_3}
\end{figure*}

Observing that Transformer-based methods have the ability to distinguish instances and inspired by the part-based method \cite{ref17}, \cite{ref18}, \cite{ref19}, \cite{ref20}, \cite{ref21}, \cite{ref22}. A new approach for Feature Segmentation and Region Alignment (FSRA) is proposed to achieve segmentation of specific instances (patch-level) and feature alignment of regions (region-level) for the purpose of extracting the corresponding parts and aligning features even when there are position deviations or scale changes between images. The proposed FSRA consists of two parts. The first one is Heatmap Segmentation Module (HSM): As shown in the light green part in the middle of Fig. \ref{fig_3}, this module divides the feature map according to the heat distribution of the feature map, and splits the feature map into several blocks from 1 to n to achieve the segmentation of patch-level instances. The other part is the Heatmap Alignment Branch (HAB): According to the segmentation feature map of HSM, the parts corresponding to different viewpoints are cut out in turn to calculate the loss, which helps the network to learn the desired heat distribution rules. As shown in the light blue part of Fig. \ref{fig_3}, where the left side is the image taken by the UAV and the right side is the satellite image, both of which are approached by HAB to close the distance of the corresponding blocks.

In addition, inspired by LCM \cite{ref24}, we realize that satellite images are highly scarce in the University-1652 [1] datasets, and expanded images can effectively improve network learning capabilities. In view of that, we propose a multiple sampling strategy to expand satellite imagery. The proposed multiple sampling strategy will increase the training time, but will not cause any additional burden on inference. Experiments show that our multiple sampling strategy can effectively improve the accuracy of the model. 

In short, the main contributions of this paper are as follows.

\begin{itemize}
\item{We propose a transformer-based strong baseline for cross-view geo-localization and achieve competitive performance with CNN-based frameworks.}
\item{For the problem caused by position offset and uncertainty of distance and scale, we designed FSRA to implement patch-level segmentation and region-level alignment.}
\item{We have carefully analyzed and improved some tricks to try to solve some problems in cross-view geo-localization. To resolve the problem of sample size imbalance under different perspectives in the University-1652, a multiple sampling strategy that increases accuracy with no pain was proposed. To further improve the performance of cross-view geo-localization, we exhaustively analyzed the impact of KLLoss \cite{ref25} and TripletLoss and made new improvements on TripletLoss.}
\item{The final framework FSRA achieves state-of-the-art performance on both tasks of drone view target localization and drone navigation in the University-1652.}
\end{itemize}

\section{RELATED WORK}
In this section, we briefly review related previous works, including \textbf{Cross-View Geo-Localization and Transformer In Vision}.

\subsection{Cross-View Geo-Localization}
Cross-view geo-localization mainly focuses on two matching tasks: the matching of ground and satellite views and the matching of drone and satellite views. CVUSA \cite{ref26} and CVACT \cite{ref27} constructed a panoramic street-view image to match the satellite-view image which is a challenging task, with a change in perspective spanning around 90 degrees. Recently, a large-scale benchmark called VIGOR \cite{ref52}, which beyond One-to-one Retrieval, was proposed to bridge the gap between the realistic setting and existing geo-localization datasets. University-1652 [1] innovatively proposed two missions based on drone perspective: drone view target localization and drone navigation, which proposed the drone-view as a transition view, reducing the difficulty of cross-view geo-localization. 

\textbf{Efficient Loss Function.} A popular pipeline for cross-view is to design suitable loss functions to train a CNN backbone, which is used to extract features from images. The CrossEntropy loss \cite{ref28}, TripletLoss \cite{ref29}, \cite{ref30}, and contrastive loss \cite{ref31} are most widely used in the task of retrieval. Zheng et al. \cite{ref32} applied instance loss and verification loss together to optimize the network, and achieve competitive results. Hu et al. \cite{ref33} proposed a weighted soft margin ranking loss, which not only speeds up the training convergence but also improves the retrieval accuracy. Luo et al. \cite{ref34} proposed a BNNeck to improve the coordination of ID loss and TripletLoss. Sun et al. \cite{ref35} proposed a unified perspective to optimize ID loss and TripletLoss.

\textbf{Part-based Fine-grained Features.} Focusing on the fine-grained information of different parts helps the model learn more comprehensive features. In addition, by dividing and supervising the feature maps, the sub-salience features in the image will be fully excavated. Fine-grained regions can be manually generated by person but also can be automatically learned by supervised methods. And the part-based fine-grained features have been proven reliable in the task of retrieval \cite{ref36}, \cite{ref37}, \cite{ref38}, \cite{ref39}, \cite{ref40}. LPN \cite{ref11} proposed the square-ring partition strategy to allow the network to pay attention to more fine-grained information at the edge and achieved a huge improvement. PCB \cite{ref17} applied a horizontal splitting method for human body parts to extract high-level segmentation features. AlignedReID++ \cite{ref21} automatically aligned slice information without introducing additional supervision to solve pedestrian misalignment problems caused by occlusion, view variation, and attitude deviation. MGN \cite{ref18} designed a slicing network that combines multi-branch and characterization metric dual learning strategies to extract global coarse-grained and local fine-grained features. MSCAN \cite{ref18} proposed Spatial Transform Networks to learn the local features of various parts of the human body, and merge the local features and global features into the final feature representation. PL-Net \cite{ref20} introduces a part loss to realize automatic detection of various parts of the human body, thereby increasing the discrimination on unseen persons. Rodrigues et al.\cite{ref53} addressed the temporal gap between scenes by proposing a semantically driven data augmentation technique that gives Siamese networks the ability to hallucinate unseen objects, and then apply a multi-scale attentive embedding network to perform matching tasks. Our proposed FSRA is also one of the part-based methods which is inspired by the LPN, the difference is that we do not add additional supervision but achieve automatic region segmentation, which makes our FSRA have excellent robustness and resistance to position shift.

\subsection{Transformer In Vision}
The attention mechanism \cite{ref41} of the transformer model was first proposed to solve problems in the field of Natural Language Processing. Subsequently, the strong visual performance of the transformer shown the superiority of its structure. Recently, Han et al. \cite{ref42} and Salman et al. \cite{ref43} investigated the application of the transformer in the field of computer vision. 

\textbf{Transformer In Various Field.} Alexey et al. \cite{ref15} first applied the transformer model to the task of classification, and then the development of the transformer in vision was in full swing. The transformer has achieved competitive results in most mainstream visual fields, such as object detection, semantic segmentation, GAN, Super-Resolution, Reid, etc. DETR \cite{ref44} was the first object detection framework that successfully integrates the transformer as the central building block of the detection pipeline. SETR \cite{ref45} treated semantic segmentation as a sequence-to-sequence prediction task through a pure transformer. TransGAN \cite{ref46} built a generator and a discriminator based on two transformer structures. TTSR \cite{ref47} restored the texture information of the image super-resolution result based on the transformer. TransReID \cite{ref23} applied the transformer to the field of retrieval for the first time and achieved similar results with the CNN-based method. Yu et al. \cite{ref59} extend transformer model to Multimodal Transformer (MT) model for image captioning and significantly outperformed the previous state-of-the-art methods. 

\textbf{Combination Of CNN And Transformer.} ConvTransformer\cite{ref54} mapped the input sequence to a feature map sequence using an encoder based on a multi-headed convolutional self-attentive layer, and then decoded the target synthetic frame from the feature map sequence using another deep network containing a multi-headed convolutional self-attentive layer. Conformer \cite{ref55} relied on Feature Coupling Unit (FCU) to interactively fuse local and global feature representations at different resolutions. Mobile-Former \cite{ref56} was a parallel design of MobileNet and Transformer with a bi-directional bridge which enabled bi-directional fusion of local and global features.

\textbf{Transformer In Cross-View.} In the cross-view domain, some novel and effective transformer structures have also been proposed to implement different downstream tasks.Chen et al. \cite{ref50} proposed a pair of cross-view transformers to transform the feature maps into the other view and introduce cross-view consistency loss on them. Yang et al. \cite{ref57} presented a novel framework that enables reconstructing a local map formed by road layout and vehicle occupancy in the bird’s-eye view given a front-view monocular image only, and a cross-view transformation module was proposed to strengthen the view transformation and scene understanding. Tulder et al. \cite{ref58}  presented a novel cross-view transformer method to transfer information between unregistered views at the level of spatial feature maps, which achieved remarkable results in field of Multi-view medical image analysis. Yang et al. \cite{ref60} proposed a simple yet effective self-cross attention mechanism to improve the quality of learned representations. Which improved the generalization ability and encourages representations to keep evolving as the network goes deeper.

\section{Proposed Method}
In this section, we will introduce the details of our proposed method, the complete network structure as shown in Fig. \ref{fig_3}. Firstly, the structure of the vision transformer will be introduced in Section III-A. Secondly, we will introduce the details of the proposed FSRA in Section III-B. Then a multiple sampling strategy to improve accuracy without pain will be introduced in Section III-C. Finally, we will introduce other tricks we applied in Section III-D, including the specific process of our implementation of TripletLoss and mutual learning.

\subsection{Transformer-Based Strong Baseline}
Following the general strong baseline for the University-1652 benchmark [1], we build a transformer-based strong baseline for cross-view geo-localization. Our baseline consists of two parts: feature extraction and classification supervised learning. As in Fig. \ref{fig_2}. Given an input $x\in{\mathbb{R}^{H\times{W\times{C}}}}$, where $H$, $W$, $C$ represent its height, width, and channels. Then input will be divided into N fixed-size patches\{$x_p^i|i=1,2,\cdots,N$\} and flatten into a sequence. An extra learnable embedding token denoted as $x_{cls}$ is merged into spatial information to extract robust features through supervision learning. The output $[cls\_token]$ as shown in Fig. \ref{fig_2} is regarded as a global feature representation $f$. Position information is added to each patch through learnable position embedding. The input sequence can finally be expressed as follows. 

\begin{equation}
\label{eq1}
\mathcal{Z}_0=[{\ x}_{cls};\mathcal{F}(x_p^1);\mathcal{F}(x_p^2);\cdots{};\mathcal{F}(x_p^N)]+\mathcal{P}
\end{equation}

Where $\mathcal{Z}_0$ represents input sequence embeddings. $\mathcal{F}$ is linear projection mapping the patches to $D$ dimensions. $\mathcal{P}\in{\mathbb{R}^{(N+1)\times{D}}}$ is the position embeddings. $L$ in Fig. \ref{fig_3} represents the depth of the transformer layers. The transformer attention mechanism allows each layer of the transformer to have insight into the global context, which overcomes the limitation of the receptive field of the convolutional neural network. In addition, the down-sampling operation is no longer needed.

\textbf{Position Embeddings.} The image classification and the cross-view tasks are different in the resolution of the input, so the position embedding parameters can not directly be loaded from the pre-training weights on ImageNet. The parameters of position embedding are learnable.

\textbf{Extra Learnable Embedding.} The characteristic of the transformer structure is that it does not change the dimensions of the input data, and the output contains contextual information, which can represent global features. An Extra learnable parameter is added to the input to act as a global feature vector, and the parameters are also learnable.

\textbf{Transformer Layers.} Transformer Layers play the same role as the backbone to extract the contextual semantic relationship between each patch. Its structure has shown on the right side of Fig. \ref{fig_2}, which takes all Patches containing Position Embedding as inputs, and finally outputs feature vectors of the same dimension as the original inputs after Multi-Head Attention.
	
\textbf{Supervision Learning.} Transformer-based strong baseline only regards classification results as supervision information and applies CrossEntropy loss without label-smooth as ID loss. 

\subsection{Concrete Implementation Of FSRA}
Experiments in \emph{Effect of the Transformer in Cross-View} show that the transformer-based strong baseline can achieve impressive performance in cross-view geo-localization. However, positional shift and Uncertainty of distance and scale are still major challenges to overcome. Although it is important to extract global features that are robust and contextually linked, much previous work has also shown that part-based methods are significantly more effective for image retrieval.

Aligning each part with features is a straightforward way to allow part-based methods to achieve end-to-end training. Based on that, we consider whether there is a reasonable and simple way for the model to learn the category to which each patch belongs, such as buildings, roads, and trees, so that we can segment and align them according to the category to which they belong. We suppose whether it is possible to cut out the characteristics of different categories according to the appearance of the heatmap and analyze the above problems as follows.

\begin{figure}[!t]
\centering
\includegraphics[width=0.49\textwidth]{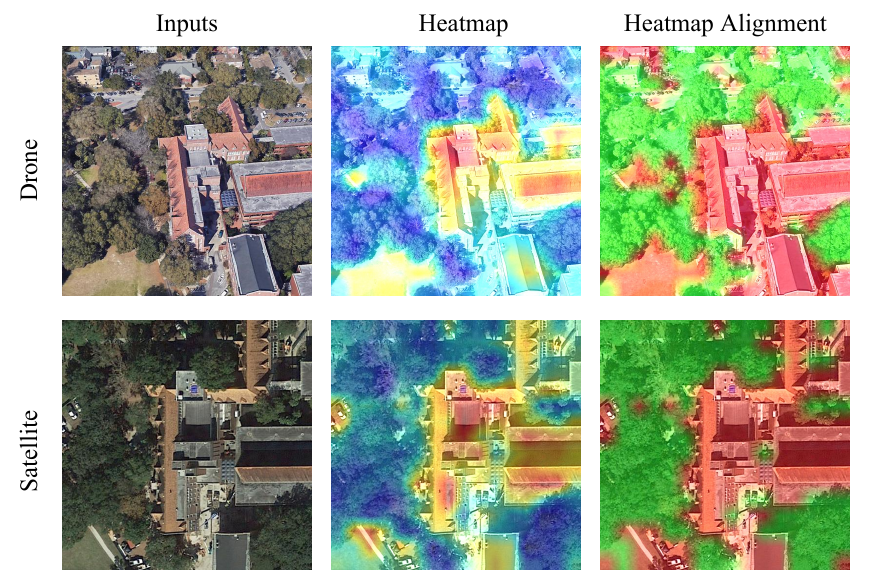}
\caption{ The left column of the figure is the input images from the drone-view and the satellite-view in the same geographic location. In the middle is the heatmap corresponding to the output of the FSRA. The right is the regional distribution generated by the HSM, the red part can be understood as the building part (foreground), and the green part is the background.}
\label{fig_4}
\end{figure}

\textbf{How to segment specific content.} HSM was proposed to achieve the purpose of segmenting different instances such as buildings, roads, and trees. The overall idea is very simple. As in Fig. \ref{fig_4}, we take $n=2$ as an example and divide the heatmap into two categories based on the magnitude of the heat value, with the large heat value being the foreground and the small heat value representing the background. As shown in the thermodynamic diagram, it is easy to see that most of the building parts have larger thermal values, while the trees and background parts have smaller thermal values. The network pays different levels of attention to different parts, which would produce a certain regularity in the distribution of heatmap. HSM is inspired by that. We perform a uniform segmentation of the feature map according to the thermal distribution. As shown on the right side of Fig. \ref{fig_4}, it is obvious that we have almost entirely distinguished the buildings from the other instances. 

In the following, we will describe the detailed implementation steps of the segmentation. Firstly, we get all the outputs  $L\in{\mathbb{R}^{B\times{N}\times{S}}}$ (where $B$ stands for batch size, $N$ stands for patch size, and $S$ stands for the length of the feature vector corresponding to each patch) except for $cls\_token$ through the forward propagation of the transformer, which can be represented as follows. 

\begin{equation}
\label{eq2}
\mathcal{L}=[\mathcal{F}(x_p^1);\mathcal{F}(x_p^2);\cdots{};\mathcal{F}(x_p^N)]
\end{equation}

The thermal value of each patch can be represented as follows.

\begin{equation}
\label{eq3}
P^c=\frac{1}{S}\textstyle\sum_{i=1}^S{\mathcal{M}^i} \ \ \ \ c=\{1,2,\cdots{},N\}
\end{equation}

where $P^c$ represents the heat value of the $c^{th}$ patch. $\mathcal{M}^i$ represents the  $c^{th}$ patch corresponds to the  $i^{th}$ value of the feature vector. In short, we do an averaging operation for the feature vector of each patch to represent the thermal value of the patch. Then, we sort the value of $P^{1-N}$ in descending order and divide patches equally according to the number of regions $n$. The number of patches corresponding to each region is as follows.

\begin{equation}
\label{eq4}
\mathcal{N}^i=\{
\begin{array}{l}\ \ \ \ \ \ \ \ 
\lfloor{}\frac{N}{n}\rfloor{}\ \ \ \ \ \ \ \ \ \ \ \ \ 
i=\{1,2,\cdots{},n-1\} \\
N-(n-1)\times{}\lfloor{}\frac{N}{n}\rfloor{}\ \ 
\ i=n
\end{array}
\end{equation}

where $\mathcal{N}^i$ represents the number of patches for the $i^{th}$ region, $\lfloor{}\cdot\rfloor{}$ is the floor function. Finally, divide $\mathcal{L}$ into $n$ parts in order, and each part corresponds to a region and then we can tag each region as a category as shown in the right column of Fig. \ref{fig_4}. Relying on HSM alone does not allow the model to move in the direction of focusing on what we want, so we need to develop an alignment supervision for this partitioning law to allow the model to distinguish between instances. The number of regions $n$ is a hyperparameter. In the following ablation experiments, we found that $n=3$ performed the best. The proposed HSM is located in the light green part of Fig. \ref{fig_3}. It is worth mentioning that HSM is implemented based on patch-level.

\begin{figure}[!t]
\centering
\includegraphics[width=0.5\textwidth]{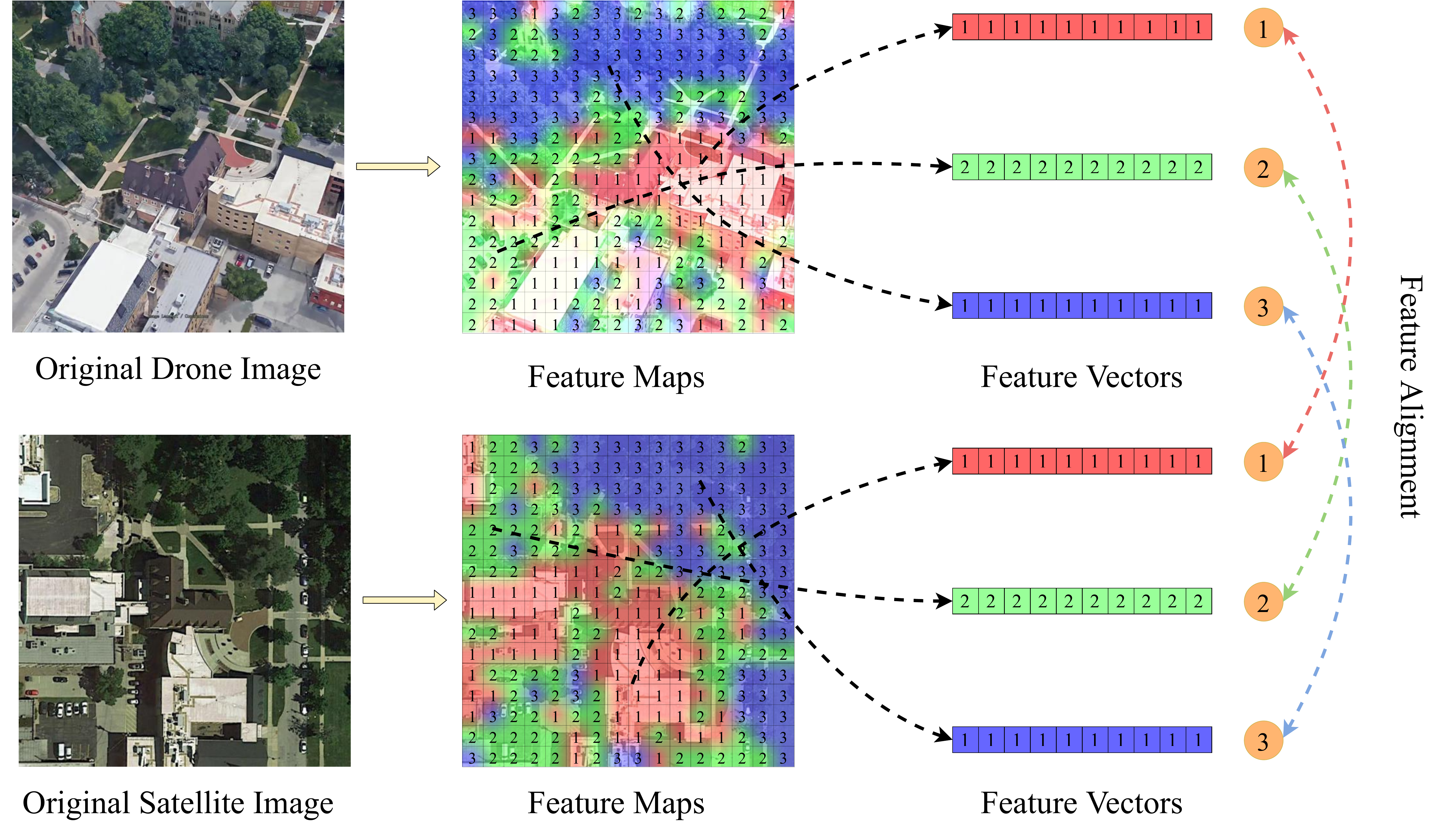}
\caption{ The left column is the input images from the drone and satellite views at the same geographic location, the middle column is the featuremaps generated by HAB with regions $n=3$. In the right is the feature vector obtained by the average pooling of each region.}
\label{fig_5}
\end{figure}

\textbf{Alignment between specific content.} HAB was proposed to achieve the effect of feature alignment. As in Fig. \ref{fig_5}. After successfully segmenting the specific content, we divide all patches into $n$ regions. Fig. \ref{fig_5} takes $n=3$ as an example. In essence, all patches are divided into 3 categories, and we use 1-3 to distinguish. The next step is to perform feature alignment based on the corresponding content in different regions. We respectively take out the part of buildings as $f_1$, the part of roads as $f_2$, and the part of trees as $f_3$. Then a pooling operation is performed on $f_{1-3}$ to obtain the feature vector $V_i\in{\mathbb{R}^{B\times{}N^i\times{}S}}, \ i=\{1,2,3\}$ that characterizes each specific content. The visualization process can be seen on the right side of Fig. \ref{fig_5}. The expression of $V_i$ is as follows.

\begin{equation}
\label{eq5}
\mathcal{V}_i=\frac{1}{\mathcal{N}^i}\textstyle\sum_{j=1}^{\mathcal{N}^i}f_i^j\ \ \ \ \ \ i=\{1,2,\cdots{},n\}
\end{equation}

where $n$ stands for the number of regions ($n$ is set to 3 in Fig. \ref{fig_5}). $f_i^j$ stands for the feature vector of the $j^{th}$ patch of the $i^{th}$ instance region. In short, $V_i$ is  obtained by taking out all the patches in each region and taking the average pooling operation.

After the above steps, we obtain the vector expression of the corresponding feature content, and then we classify each feature content separately through a $Classifier Layer$. In addition, to allow the model to establish more accurate matching relationships, we apply TripletLoss as in Fig. \ref{fig_3} to all regions to narrow the distance between regions. The specific implementation will be explained in section III.D. The proposed HAB is located in the light blue part of Fig. \ref{fig_3}. 

It is worth noting that our HAB method is region-level feature alignment, and the division of regions is determined by HSM. The reason why HAB can achieve good performance is that it distinguishes the features of different instances, which is conducive to the model not only paying attention to the global salient features, but also paying attention to the details of the background, which will help the model extraction more comprehensive fine-grained features. 

\subsection{A Multiple Sampling Strategy}

There are some unstable factors during the training process based on the transformer model.  For example, a model with the same settings is trained twice, the results obtained will have a large margin. However, We found that the main reason may be that there is only one image per category in the satellite-view, which results in only one image from other views at one time. This case will cause an imbalance between the satellite images and other images. Therefore, a multiple sampling strategy is proposed to alleviate the problem of sample imbalance.

We set a hyperparameter $k$, which represents the number of sampling. The specific implementation is as follows. Firstly, derive the image under the satellite perspective from the Unversity-1652, and enhance it to generate $k$ augmented satellite images. Augmentation methods include random shifting, random padding, random cutting, random color enhancement, etc. At the same time, $k$ images from other perspectives are randomly selected, which is the same category as the corresponding satellite perspective. 

The detailed experiment on the number of sampling $k$
was conducted in the part of the ablation study in Section IV, and the results of the experiment show that FSRA performed best when $k=3$.

\subsection{Other Tricks On Cross-View}

\textbf{Mutual Learning Based on Cross-View.} Cross-view geo-localization is a multi-input and multi-output task. Given this, we introduce a method of self-distillation. The specific implementation is as follows. Establish learning relationships between outputs from different domains to narrow the distance between similar instances. The calculation formula of KL divergence loss is shown below.

\begin{equation}
\label{eq6}
KLDiv(O_1\vert{}\vert{}O_2)=\sum_{i=1}^Np(O_1^i)\cdot{}log\frac{p(O_1^i)}{q(O_2^i)}
\end{equation}

\begin{equation}
\label{eq7}
 p(x_i)=log(\frac{e^{x_i}}{\sum_je^{x_j}}\ )
\end{equation}

\begin{equation}
\label{eq8}
q(x_i)=\frac{e^{x_i}}{\sum_je^{x_j}}\
\end{equation}

where $O_1$ represents the target output. $O_2$ represents the learning output from the model. The mutual learning loss function is expressed as follows. 

\begin{equation}
\label{eq9}
KLLoss=KLDiv(O_d\Vert{}O_s)+KLDiv(O_s\Vert{}O_d)
\end{equation}

where $O_d$ stands for the output of the drone-view image after forwarding propagation. $O_s$ stands for the output of the satellite-view image after forwarding propagation.

We verify the effectiveness of KLLoss in the ablation study. Experiments show that when KLLoss is applied alone, the accuracy of the model is significantly improved, but when KLLoss is applied together with TripletLoss, the accuracy of the model is not improved significantly. This may be caused by the same optimization direction of TripletLoss and KLLoss. 

\begin{figure}[!t]
\centering
\includegraphics[width=0.5\textwidth]{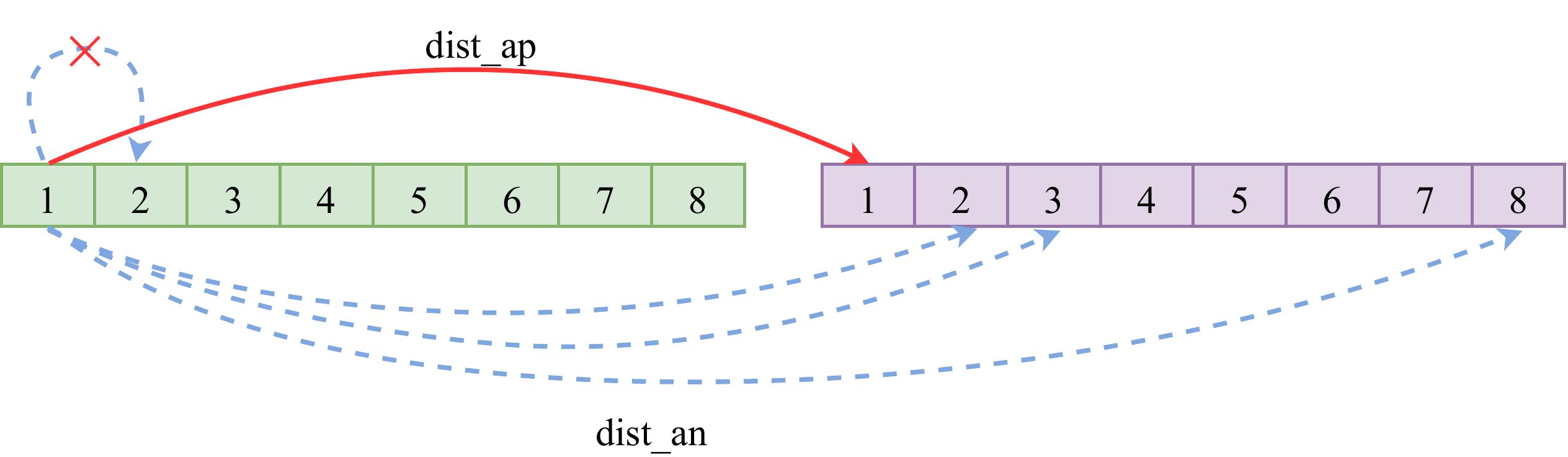}
\caption{The number 1-8 indicates the category the image belongs to. The light green part represents 8 images from the views of drones or satellites, the light purple part represents 8 images from the views of satellites or drones, dist\_ap represents the distance between pictures of the same category, and dist\_an represents the distance between pictures of different categories, The red × stands for that the distance is not calculated for images from the same views.}
\label{fig_6}
\end{figure}

\textbf{TripletLoss based on Cross-View.} Only using CrossEntropy loss can not make the model end-to-end. When testing the accuracy of the model, Euclidean distance is used to judge the similarity between samples. TripletLoss can act as a supervisor to narrow the distance between the same targets from different domains. The TripletLoss can be formulated as follows.
\begin{equation}
\label{eq10}
TL={\lVert{}d(a,p)-d(a,n)+M\rVert{}}_+
\end{equation}

\begin{equation}
\label{eq11}
d(a,x)={\lVert{}a-x\lVert{}}_2\
\end{equation}

where $\lVert{}\cdot\rVert{}_{+}$ represents max($\cdot$,0) operation. $\lVert{}\cdot\rVert{}_{2}$ represents a 2-norm operation. $M$ is the value of margin. We apply Euclidean distance in Equation \ref{eq11} to measure the distance between vectors. In Equation \ref{eq10}, we compute the TripletLoss with $M=0.3$ in all our experiments.

Unlike traditional TripletLoss, the task of cross-view is to match images from different domains, and it is not essential to be distinguished from images of the same perspective. Therefore, we only calculate TripletLoss for images between different views. As in Fig. \ref{fig_6}, for example, we take out an image from the light green set (drone/satellite view) to calculate the TripletLoss with all images from the light purple set (satellite/drone view).

\begin{table}[h]
\renewcommand\arraystretch{1.4}
\caption{Statistics the number of images, buildings, and universities from view of drone, satellite, and street in the training set and test set of the University-1652 dataset. And statistics the images number of query and gallery in the test set. There are no duplicate universities in the training set and test set.}
\label{table1}
\resizebox{1.0\hsize}{!}
{
\begin{tabular}{cc|c|c|c|c}
\specialrule{0.75pt}{0pt}{0pt}

\multicolumn{2}{c|}{split} & {views} & {images} & {classes} & university \\
\specialrule{0.5pt}{0pt}{0pt}
\multicolumn{2}{c|}{\multirow{3}{*}{\rotatebox{90}{Train}}} & {Drone} & {37,854} & {701} & \multirow{3}{*}{33} 
\\
& &{Satellite}&{701}&{701}&
\\
& &{Street}&{11,640}&{701}&
\\
\specialrule{0.25pt}{0pt}{0pt}
\multirow{6}{*}{\rotatebox{90}{Test}}& \multirow{3}{*}{\rotatebox{90}{Query}}&{Drone}&{37,855}&{701}&\multirow{6}{*}{39}
\\
& &{Satellite}&{701}&{701}&
\\
& & {Street} & {2,579} & {701} & 
\\
& \multirow{3}{*}{\rotatebox{90}{Gallery}}  & {Drone} & {51,355} & {951} &  
\\
& & {Satellite} & {951} & {951} & \\
& & {Street} & {2,921} & {793} &
\\
\specialrule{0.75pt}{0pt}{0pt}
\end{tabular}
}
\end{table}

\section{Experiment}
We first introduce a large-scale cross-view geo-localization dataset in Section IV-A. Then Section IV-B describes the implementation details. We provide the comparison with state-of-the-art methods in Section IV-C, followed by the ablation study in Section IV-D. 

\subsection{Datasets And Evaluation Protocol}

Our method is mainly used to solve UAV-related problems, including drone view target localization and drone navigation. We have done a lot of experiments based on the large-scale dataset, University-1652 [1]. Table I shows the number of images from different views of the University-1652 dataset during training and testing. The column of classes indicates the number of buildings, and the column of university indicates the number of universities included in the sample. The entire dataset contains a total of 72 universities, and there is no intersection between the training set and the test set.

\textbf{University-1652} is a multi-view multi-source benchmark for drone-based geo-localization, which contains images from three platforms, i.e., synthetic drones, satellites, and ground cameras. University-1652 is the first large-scale geo-localization dataset contained drone-view and enables two tasks, i.e., drone view target localization (Drone $\rightarrow$ Satellite) and drone navigation (Satellite $\rightarrow$ Drone). It aims to improve the accuracy of matching the images between drone-view and satellite-view. The dataset collected 1,652 buildings from 72 universities in the world. As in Table I, the training set includes 701 buildings of 33 universities, and the testing set includes the 951 buildings of the 39 universities. The buildings in the training set and the test set have no overlap. There are 701 buildings with 50,195 images for training, which contains 37,854 drone-view images, 701 satellite-view images, and 11,640 street-view images. 

For testing, In the drone view target localization task (Drone$\rightarrow$Satellite), there are 37,855 drone-view images in the query set and 701 true-matched satellite-view images, and 250 satellite-view distractors in the gallery. There is only one true-matched satellite-view image under this setting. In the drone navigation task (Satellite → Drone), there are 701 satellite-view query images, and 37,855 true-matched drone-view images, and 13,500 drone-view distractors in the gallery. There are about 54 true-matched drone-view images that can be matched.

\subsection{Implementation Details}

\textbf{In data processing.} We apply a multiple sampling strategy. Considering that there is only one satellite image for each category, image augmentation is applied to extend the satellite set for alleviating the imbalance of images in different domains.

\textbf{In network structure and training strategy.} We adopt a small size Vision Transformer (Vit-S) pretrained on ImageNet as our backbone. We have adopted the FSRA structure which regions the output of the transformer by HSM, and aligns the feature map by HAB. In terms of parameter initialization, we adopt kaiming initialization \cite{ref49} for the classifier module. In training, we resize the input image to the size of $256\times{}256$ and perform image augmentation, e.g., random padding, random cropping, and random flipping. For the optimizer, we adopt stochastic gradient descent (SGD) with momentum 0.9 and weight decay 0.0005 with a mini-batch of 8. For the setting of the initial learning rate, the backbone parameter is set to 0.003, and the rest of the learnable parameters are set to 0.01. The learning rate of all parameters are decayed by 0.1 in the epoch of 70 and 110 respectively, the model is trained for 120 epochs in total. 

\textbf{In the loss function.} We use the CrossEntropy loss as the classification loss function and adopt TripletLoss with a margin of 0.3 to narrow the distance of the same target from different domains. Besides, KL divergence loss is introduced to narrow the distance of the classification vectors. 

\textbf{During the test.} We utilize the Euclidean distance to calculate the similarity between query images and candidate images in the gallery set. Our model is based on the framework of Pytorch, and all experiments are performed on Nvidia GTX 1080Ti GPU.

\begin{table*}[h]
\tiny
\renewcommand\arraystretch{1.2}
\caption{Comparision with state-of-the-art results which have reported in University-1652. M represents the margin of TripletLoss, k represents the number of sampling, s represents the size of input images and Vit-S represents the Small-scale vision transformer network..}
\label{table2}
\resizebox{0.95\hsize}{!}
{
\begin{tabular}{c|c|cc|cc}
\specialrule{0.5pt}{0pt}{0pt}   

\multirow{2}{*}{Method} & 
\multirow{2}{*}{Backbone} &
\multicolumn{2}{c|}{$\ \ \ \ $Drone$\rightarrow$Satellite$\ \ \ \ $}& 
\multicolumn{2}{c}{$\ \ \ \ $Satellite$\rightarrow$Drone$\ \ \ \ $}\\
 & &{$\ \ \ $R@1}&{$\ \ \ $AP}&{$\ \ \ $R@1}&{$\ \ \ $AP}
\\
\specialrule{0.25pt}{0pt}{0pt}
{Contrastive Loss\cite{ref31}}&{VGG16}&{$\ \ \ $52.39}&{$\ \ \ $57.44}&{$\ \ \ $63.91}&{$\ \ \ $52.24}
\\
{Weighted Soft Margin TripletLoss \cite{ref33}}&{VGG16}&{$\ \ \ $53.21}&{$\ \ \ $58.03}&{$\ \ \ $65.62}&{$\ \ \ $54.47}
\\
{TripletLoss (M = 0.3) \cite{ref29}}&{ResNet-50}&{$\ \ \ $55.18}&{$\ \ \ $59.97}&{$\ \ \ $63.62}&{$\ \ \ $53.85}
\\
{Instance Loss + GeM Pooling \cite{ref48}}&{ResNet-50}&{$\ \ \ $65.32}&{$\ \ \ $69.61}&{$\ \ \ $79.03}&{$\ \ \ $65.35}
\\
{Instance Loss [1]}&{ResNet-50}&{$\ \ \ $58.23}&{$\ \ \ $62.91}&{$\ \ \ $74.47}&{$\ \ \ $59.45}
\\
{LCM (ResNet-50) \cite{ref24}}&{ResNet-50}&{$\ \ \ $66.65}&{$\ \ \ $70.82}&{$\ \ \ $79.89}&{$\ \ \ $65.38}
\\
{LPN \cite{ref11}}&{ResNet-50}&{$\ \ \ $75.93}&{$\ \ \ $79.14}&{$\ \ \ $86.45}&{$\ \ \ $74.79}
\\
\specialrule{0.25pt}{0pt}{0pt}
{Ours (k=1)}&{Vit-S}&{$\ \ \ $82.25}&{$\ \ \ $84.82}&{$\ \ \ $87.87}&{$\ \ \ $81.53}
\\
{Ours (k=3)}&{Vit-S}&{$\ \ \ $84.51}&{$\ \ \ $86.71}&{$\ \ \ $88.45}&{$\ \ \ $83.37}
\\
{Ours (k=1, s=512)}&{Vit-S}&{$\ \ \ $\textbf{85.50}}&{$\ \ \ $\textbf{87.53}}&{$\ \ \ $\textbf{89.73}}&{$\ \ \ $\textbf{84.94}}
\\

\specialrule{0.5pt}{0pt}{0pt}   
\end{tabular}
}
\end{table*}

\subsection{Comparison With Existing Methods}
On the University-1652 [1] dataset, we employ the proposed FSRA to compare with existing competitive methods. As shown in Table II, in the task of Drone $\rightarrow$ Satellite, the proposed HAB achieved 82.25\% Recall@1 and 84.82\% AP; In the task of Satellite $\rightarrow$ Drone, FSRA has achieved 88.45\% Recall@1 and 83.37\% AP. All our experiments only use drone and satellite views for training. The performance has surpassed state-of-the-art method e.g., LPN by a large margin of about 6\% AP improvement. When we adopt different sampling strategies, the experimental results of our method have been further improved. When we use $3\times$ sampling, the value of Recall@1 rises from 82.25\% to 84.51\% and the value of AP rises from 84.82\% to 86.71\% in the drone view target localization task (Drone$\rightarrow$Satellite). The value of Recall@1 rises from 87.87\% to 88.45\% and the value of AP is from 81.53\% to 83.37\% in the drone navigation task (Satellite$\rightarrow$Drone).

\subsection{Ablation Studies}

To verify the effectiveness of our method, we design several ablation experiments.

\begin{table}[h]
\renewcommand\arraystretch{1.7}
\caption{Comparison of ResNet and Vision Transformer. Inference time is measured compared to ResNet-50, and other backbones are evaluated relative to the baseline of ResNet-50. All results are performed on the same device for a fair comparison. Vit-S/16 is regarded as the baseline model and abbreviated as Baseline in the rest of this paper. Vit-B/16 is the standard model proposed in the original paper \cite{ref15}}
\label{table3}
\resizebox{1.0\hsize}{!}
{
\begin{tabular}{c c|c c c c}
\specialrule{1pt}{0pt}{0pt}   

\multirow{2}{*}{Backbone}& 
\multirow{2}{*}{Inference Time}&
\multicolumn{2}{c}{ Drone$\rightarrow$Satellite }& 
\multicolumn{2}{c}{ Satellite$\rightarrow$Drone }\\
 & &{R@1}&{AP}&{R@1}&{AP}
\\
\specialrule{0.5pt}{0pt}{0pt}   
{ResNet-50}&{1x}&{60.93}&{65.31}&{75.61}&{61.69}
\\
{ResNet-101}&{1.48x}&{65.33}&{68.32}&{79.44}&{65.43}
\\
{Vit-S/16}&\textbf{1.21x}&\textbf{71.04}&\textbf{74.62}&\textbf{83.31}&\textbf{72.08}
\\
{Vit-B/16}&{1.79x}&{73.32}&{76.88}&{84.74}&{74.72}
\\

\specialrule{1pt}{0pt}{0pt}   
\end{tabular}
}
\end{table}

\begin{figure}[!t]
\centering
\includegraphics[width=0.5\textwidth]{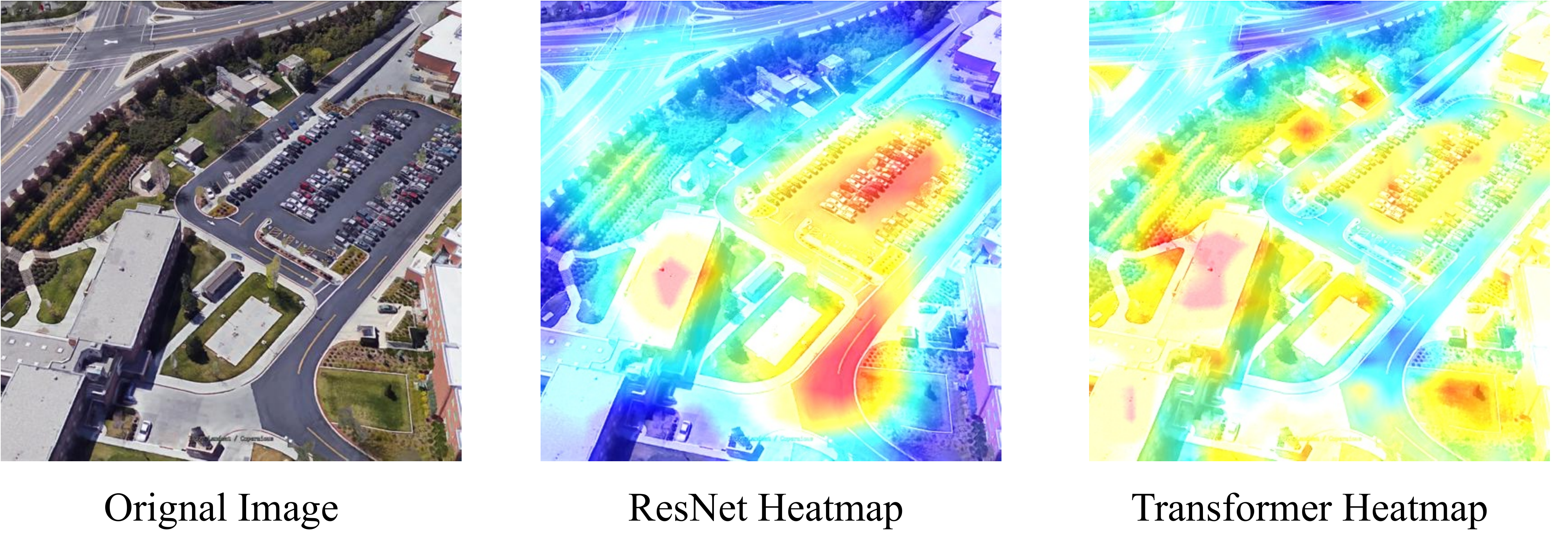}
\caption{The left is the original image, the middle is the heatmap of the last layer of ResNet-50, and the right is the heatmap of the last layer of Vit-S.}
\label{fig_7}
\end{figure}

\textbf{Effect of the Transformer in Cross-View.} We bring the transformer network structure into the field of cross-view and compare the performance of Transformer-based and ResNet-based networks. As shown in Table III. The Vit-S network with a single branch outperforms ResNet-50 by 9.31\% and outperforms ResNet-101 by 6.3\%, and the inference time is only 1.21× of ResNet-50, which is faster than ResNet-101. Besides, we also compared the accuracy and  speed of Vit-B with other backbones. We found that deepening the transformer can not bring a significant improvement. The transformer's attention mechanism has its limitations, and the impact of its model size on performance depends on the size of the data volume. University-1652 is a 10,000-level data set, which is not suitable for large-scale transformer networks. Therefore, we use Vit-S as the backbone for other ablation experiments. Through the comparison of baseline between CNN-based method and Transformer-based method, we found that there is a large margin between the CNN-based method and the Transformer-based method. Since we checked the heatmap based on ResNet-50 and Vit-S respectively. As in Fig. \ref{fig_7}. The attention mechanism allows the network to focus on global information, while the CNN-based approach will only focus on notable information but ignore the peripheral features. In addition, the heatmaps generated by the Transformer-based method can segment buildings, roads, and trees, which pave the way for our method.

\begin{figure}[!t]
\centering
\includegraphics[width=0.5\textwidth]{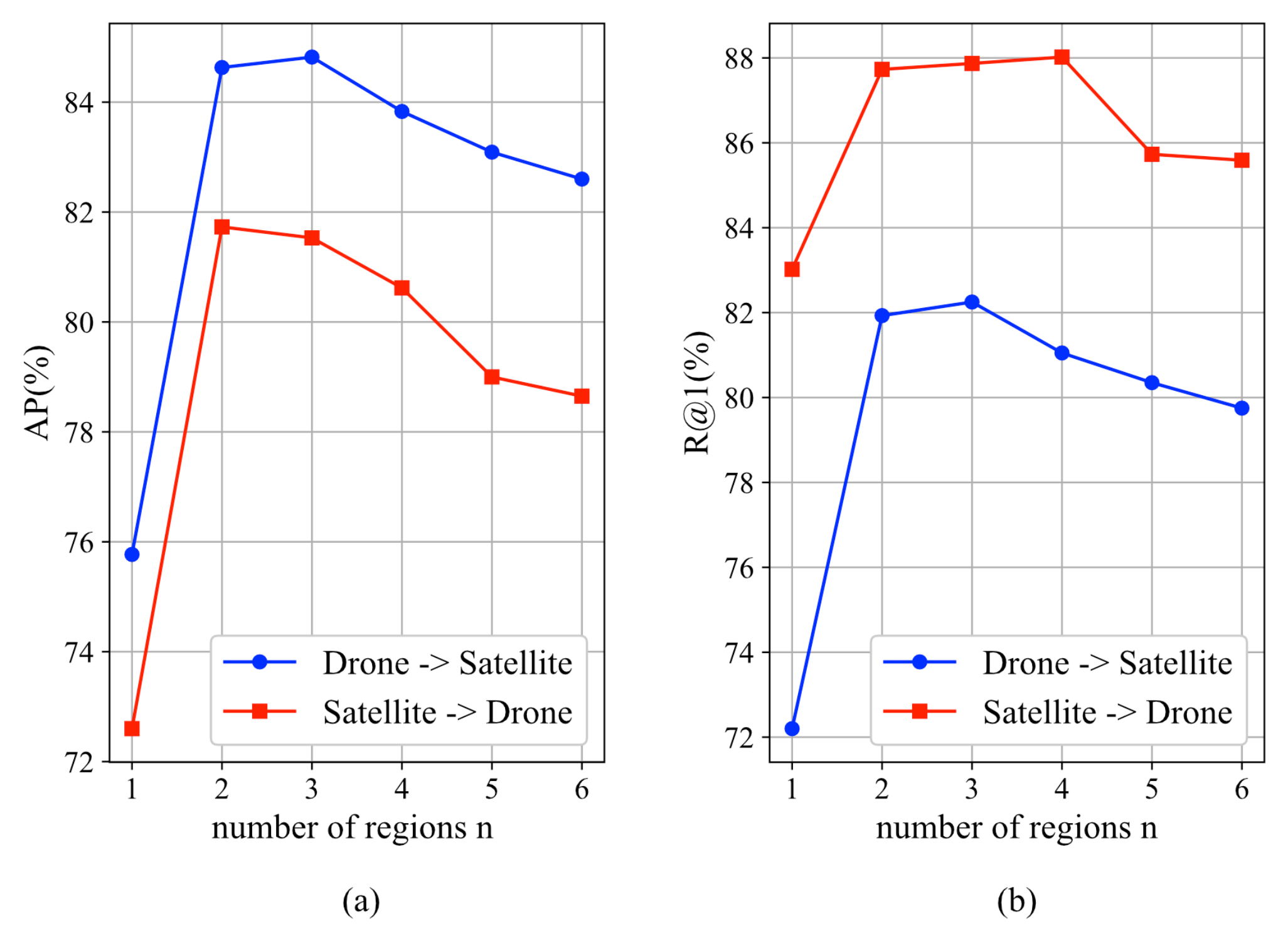}
\caption{Compare the effects of the number of regions n on the task of drone view target localization and the task of drone navigation. The red line represents the task of drone view target localization (Drone → Satellite), and the blue line represents the task of drone navigation (Satellite → Drone). Our experiments are all based on TripletLoss (M=0.3). (a) Show the effect of the number of regions n on the accuracy of Recall@1. (b) Show the effect of the number of regions n on the accuracy of AP. We find that R@1 and AP achieve the best performance when n=3.}
\label{fig_8}
\end{figure}
\textbf{Effect of the number of regions.} The number of regions is an important indicator in our network. By default, we deploy $n=3$. The model only applies the global branch of Vit when $n=0$.  When $n=1$, the HAB deploys global average pooling of the feature vectors and concats them with the global branch of Vit. We make an experiment to verify the influence of the number of regions on the accuracy of Recall@1 and AP, as in Fig. \ref{fig_8}. When the number of regions $n=3$, all indicators are the best. We believe that when $n=3$, the proposed FSRA divides the images of the University-1652 [1] dataset into three categories: buildings, roads, and trees. And the features between different domains can be well segmented and aligned. When the number of regions $n=2$, the proposed FSRA divides the image into two categories: architecture and background, which also achieves good performance. 

\begin{table}[h]
\renewcommand\arraystretch{1.5}
\caption{In the two cases of Black Pad and Flip Pad, the proposed FSRA and State-Of-The-Art method LPN correspond to the AP accuracy values of different pad sizes and the speed of decline.}
\label{table4}
\resizebox{1.0\hsize}{!}{
\begin{tabular}{c|cc|cc}
\specialrule{1pt}{0pt}{0pt}   

\multirow{2}{*}{\shortstack{Pad\\Pixel} } & \multicolumn{2}{c|}{Black Pad AP (\%)} & \multicolumn{2}{c}{Flip Pad AP (\%)} \\
 {}&{FSRA}&{LPN}&{FSRA}&{LPN}\\
\specialrule{0.5pt}{0pt}{0pt}   
{0}&\bm{$84.77_{-0}$}&\bm{$81.17_{-0}$}&\bm{$84.77_{-0}$}&\bm{$81.17_{-0}$} \\
{10}&{$84.13_{-0.64}$}&{$80.79_{-0.38}$}&{$84.19_{-0.58}$}&{$80.07_{-1.10}$} \\
{20}&{$82.7_{-2.07}$}&{$78.29_{-2.88}$}&{$82.26_{-2.51}$}&{$77.18_{-3.99}$}\\
{30}&{$80.03_{-4.74}$}&{$74.01_{-7.16}$}&{$78.46_{-6.31}$}&{$72.67_{-8.50}$}\\
{40}&{$76.41_{-8.36}$}&{$68.06_{-13.08}$}&{$73.13_{-11.64}$}&{$65.83_{-15.34}$}\\
{50}&{$71.6_{-13.17}$}&{$60.61_{-20.56}$}&{$66.07_{-18.70}$}&{$58.17_{-23.00}$}\\
{60}&\bm{$65.76_{-19.01}$}&\bm{$52.09_{-29.08}$}&\bm{$57.96_{-26.81}$}&\bm{$49.88_{-31.29}$}\\
\specialrule{1pt}{0pt}{0pt}   
\end{tabular}
}
\end{table}

\begin{figure}[!t]
\centering
\includegraphics[width=0.5\textwidth]{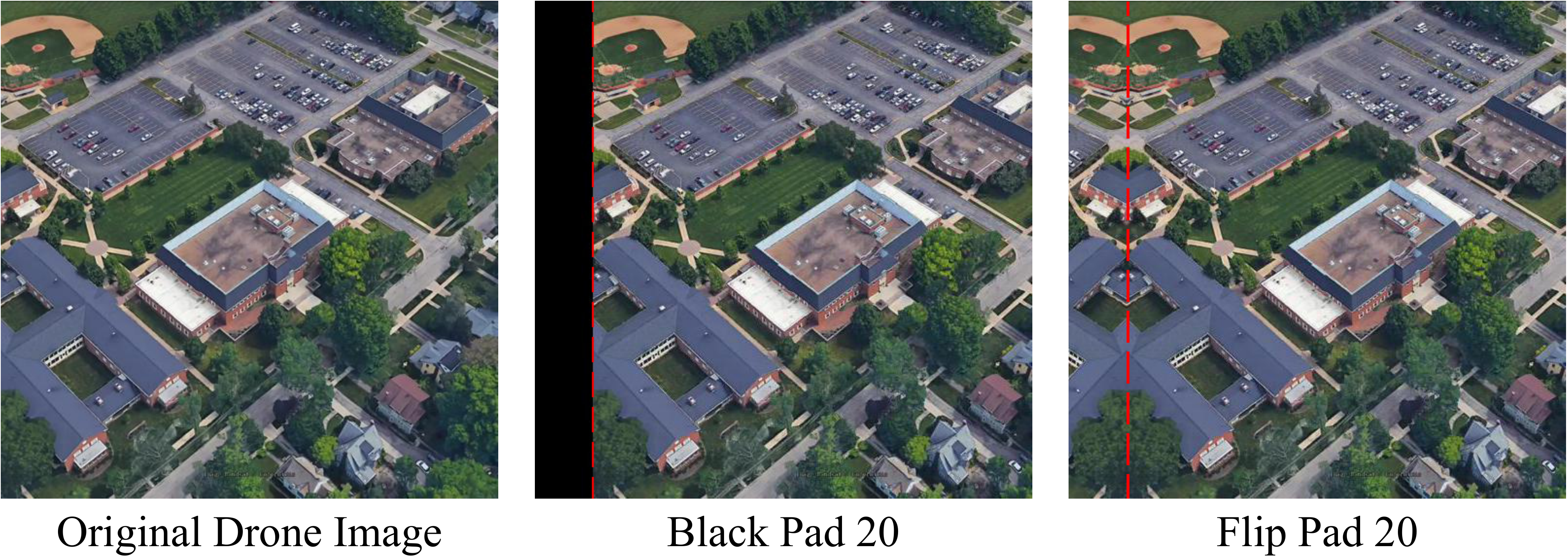}
\caption{The image on the left is the original drone image, and the middle image is the image with a width of 20 expanded with black on the left side of the image and cropped to the same width on the right side of the image. The image on the right is obtained by mirroring and expanding a 20-pixel wide portion of the left side of the image and cutting off an equal pixel width on the right side of the image. The red dotted line is the dividing line of Padding. }
\label{fig_9}
\end{figure}

\begin{figure}[!t]
\centering
\includegraphics[width=0.5\textwidth]{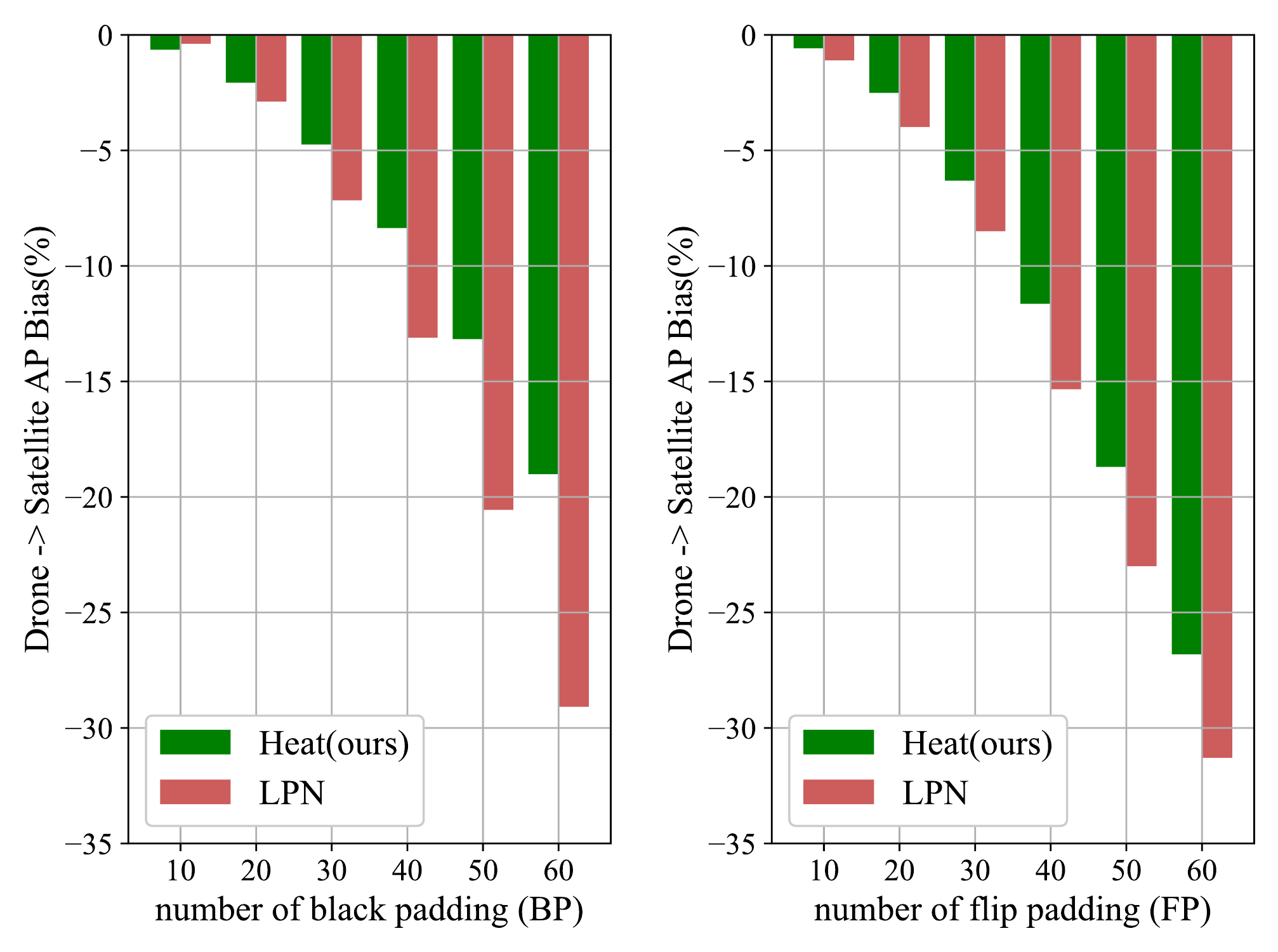}
\caption{Just like the two padding methods shown in Fig. \ref{fig_9}, we explore the impact of the number of the black pad and the flip pad on AP and Recall@1. The vertical axis represents the magnitude of the decrease in accuracy.}
\label{fig_10}
\end{figure}

\textbf{Robustness of FSRA to position shifting.} In order to verify the robustness of FSRA against position shifting, two different shifting methods are proposed for testing: $Black Pad (BP)$ and $Flip Pad (FP)$. $Black Pad$ fills the black block with width $P$ on the left side of the image and cuts out the image with width $P$ on the right side. $Flip Pad$ flips the part with width $P$ on the left side of the image and cuts out the image with width P on the right. As in Fig. \ref{fig_9}. In order to verify the anti-offset of FSRA, we compare the proposed FSRA with the state-of-the-art method LPN. As shown in Fig. \ref{fig_10}, when the padding size increases, the accuracy of FSRA decreases much slower than LPN. Besides, the accuracy of $Black Pad$ decreases more slowly than $Flip Pad$, which may be caused by the fact that $FP$ increases the confusion information at the edge and causes the uneven content distribution. As shown in Table IV, when $BP=60$, The AP of LPN was reduced by 29.08\%, while the AP of our FSRA was reduced by 19.01\%, which is about 10 points less than that of LPN. When $FP=60$, The AP of LPN was reduced by 31.29\%, while the AP of our FSRA was reduced by 26.81\%, which is about 4.5 points less than that of LPN..

The advantage of FSRA over part-based like LPN in resisting position shift mainly lies in the fact that FSRA does not artificially design regions, but allows the model to learn to a set of division rules by itself, and this segmentation is patch-level. Therefore, when the input image has a large position offset, the network can still distinguish which parts are buildings and which parts are trees. In contrast, artificially designed segmentation no longer makes sense when significant offsets occur, but is often effective in the absence of offsets and anomalies. This idea can also be applied to the field of ReID. For example, During object detection, there might be incomplete cuts of the human body, or the cut image contains a lot of background. In this case, our FSRA can still be recognized effectively by automatic segmentation.

\begin{figure}[!t]
\centering
\includegraphics[width=0.49\textwidth]{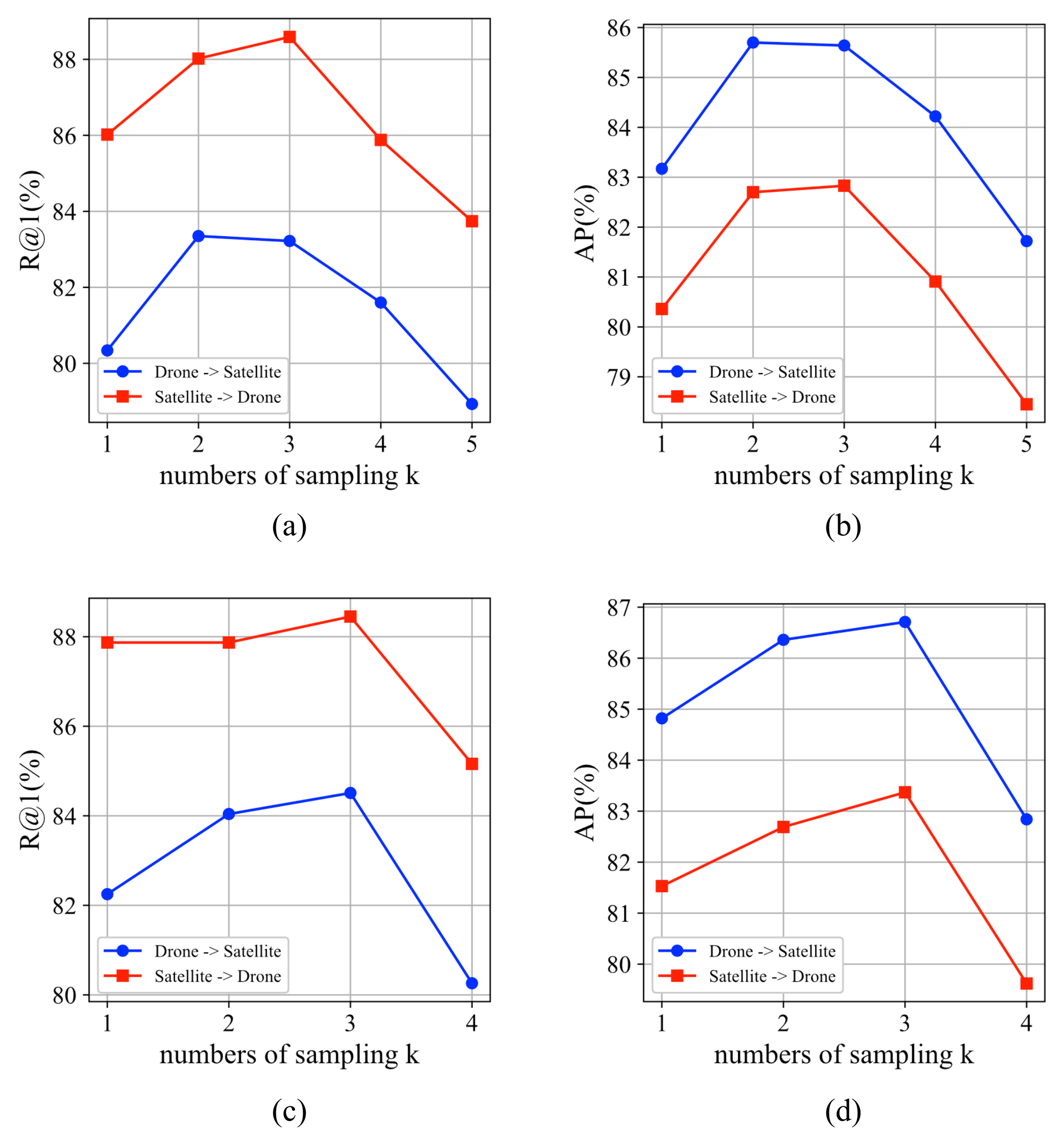}
\caption{We conducted experiments on the impact of the number of sampling on AP and Recall@1 in tasks of Drone$\rightarrow$Satellite and Satellite$\rightarrow$Drone. $k$ stands for the number of the sampling. When the number of sampling is 3, the accuracy of AP and R@1 in both tasks reaches the best.}
\label{fig_11}
\end{figure}

\textbf{The impact of sampling on accuracy.} Adequate sampling has a great influence on the fitting of the network. Unbalanced or insufficient data will lead to unstable model training, and the final results will be unsatisfactory. In the University-1652 dataset, one satellite image corresponds to 27 drone-view images. Previously, single-fold sampling was done by taking one from a specific category in each iteration, i.e., one of the 27 drone-view images and only one of the satellite-view.The multiple sampling approach can optimize two aspects of the problem: 1) the sample imbalance problem of different viewpoint images.  This problem has essentially been raised in LCM \cite{ref24} (the authors achieved the best using equal-multiplicity sampling of UAV and satellite images). 2) The number and proportion of positive and negative samples for TripletLoss. When we change the sampling multiplicity, the number of positive samples in a single batch of TripletLoss will change, which has an impact on the metric learning. To verify that our approach is not due to the effect of TripletLoss, we conducted experiments using the FSRA with region $n=3$ and no TripletLoss. As shown in Fig. \ref{fig_11}(a) and (b), the trend is up and then down in both AP and R@1 indicators, and the overall optimum is reached at $k=3$. In addition, we train the model with the addition of TripletLoss, as shown in Fig. \ref{fig_11}(c) and (d), which also shows the same trend of rising then falling and optimal at $k=3$. The parameter $k$ can be interpreted as a hyperparameter, and $k=3$ is a more effective value in Unversity-1652. $k$ affects the training time, but has no effect on the inference phase. We believe that the reason why $k$ is too large for model training is overfitting on one hand, and on the other hand, as $k$ increases, the proportion of similar samples in a single batch will increase, and the model will learn fewer inter-class differences in a single batch.We conjecture that batchsize has an impact on the choice of $k$ values in the multiple sampling strategy, which we discuss in \emph{Appendix B}.

\begin{table}[h]
\tiny
\renewcommand\arraystretch{1.4}
\caption{Ablation study on the impact of different input sizes on University-1652. The experimental results are based on the number of sampling k=1, triple loss with margin=0.3.}
\label{table5}
\resizebox{1.0\hsize}{!}{
\begin{tabular}{c|cc|cc}
\specialrule{0.75pt}{0pt}{0pt}   

\multirow{2}{*}{Image Size} & \multicolumn{2}{c|}{$\ \ \ \ \ $Drone$\ \rightarrow$Satellite\ $\ \ \ $} &
\multicolumn{2}{c}{$\ \ \ \ \ $Drone$\ \rightarrow$Satellite\ $\ \ \ $} 
\\
 &{$\ \ \ $R@1}&{$\ \ \ $AP}&{$\ \ \ $R@1}&{$\ \ \ $AP}
 \\
\specialrule{0.5pt}{0pt}{0pt}   
224 & {$\ \ \ $80.81} & {$\ \ \ $83.65} & {$\ \ \ $87.73} & {$\ \ \ $80.02} \\
256 & {$\ \ \ $82.25} & {$\ \ \ $84.82} & {$\ \ \ $87.87} & {$\ \ \ $81.53} \\
320 & {$\ \ \ $84.08} & {$\ \ \ $86.38} & {$\ \ \ $87.87} & {$\ \ \ $82.63} \\
384 & {$\ \ \ $84.82} & {$\ \ \ $87.03} & {$\ \ \ $87.59} & {$\ \ \ $83.37} \\
512 & {$\ \ \ $\textbf{85.5}} & {$\ \ \ $\textbf{87.53}} & {$\ \ \ $\textbf{89.73}} & {$\ \ \ $\textbf{84.94}} \\
\specialrule{0.75pt}{0pt}{0pt}
\end{tabular}
}
\end{table}

\textbf{Effect of the input image size.} Image with small size will compress the fine-grained information and damage the complete features of the original image. Large-scale images can often achieve higher accuracy because they maintain the original fine-grained information. In contrast, large-scale images often require larger memory resources and longer inference time during training and testing. To balance the input image size with memory usage, we conduct experiments on FSRA with the number of regions $n=3$. According to different input sizes, the experimental results are shown in Table V. In both tasks, i.e., Drone$\rightarrow$Satellite and Satellite$\rightarrow$Drone, we observe that the performance gradually improves when the input image size increases from 224 to 512, and the AP has a big improvement when the image input size is changed from 256 to 320. We hope that when the hardware resources are limited, this ablation experiment can play a reference role in selecting the appropriate input image size.

\begin{table}[h]
\renewcommand\arraystretch{1.5}
\caption{Ablation study to verify the robustness of the proposed FSRA at different distances between drones and target in University-1652.}
\label{table6}
\resizebox{1.0\hsize}{!}{
\begin{tabular}{c|cc}
\specialrule{0.75pt}{0pt}{0pt}   

\multirow{2}{*}{$\ \ \ $Distance$\ \ \ $} & \multicolumn{2}{c}{$\ \ \ \ \ \ \ \ \ $ Drone $\ \ \rightarrow \ \ $ Satellite $\ \ \ \ \ \ \ \ \ \ $} \\
 &{$\ \ \ \ \ \ $ R@1}&{$\ \ \ \ \ \ $AP}\\
\specialrule{0.5pt}{0pt}{0pt}   
{ALL} &{$\ \ \ \ \ \ $82.25}&{$\ \ \ \ \ \ $84.82} \\
{Long} &{$\ \ \ \ \ \ $79.71}&{$\ \ \ \ \ \ $82.69} \\
{Middle} &{$\ \ \ \ \ \ $84.05}&{$\ \ \ \ \ \ $86.36} \\
{Short} &{$\ \ \ \ \ \ $82.87}&{$\ \ \ \ \ \ $85.25} \\
\specialrule{0.75pt}{0pt}{0pt}
\end{tabular}
}
\end{table}

\textbf{Effect of the drone distance to the geographic target.} The scale of the satellite-view image in University-1652 is fixed, while the scale of the drone-view image changes dynamically with the distance of the drone to the geographic target. According to the distance between the drone and the target building, we divide the University-1652 dataset into three parts: \textbf{Long, Middle, and Short.}  We verify the effect of the proposed FSRA under three different levels of distance, as shown in Table VI. The proposed FSRA does not have a big margin at a different level of distance. It has the lowest accuracy at Long distances and the highest accuracy at Middle. Compared with the current state-of-the-art network e.g., LPN, which has a margin of 20\% Recall@1 and 17\% AP between Long and Middle distance, the proposed FSRA has better scale adaptive capabilities. 

\begin{table}[h]
\renewcommand\arraystretch{1.5}
\caption{Ablation studies to verify the effects of some other tricks, including KLLoss, TripletLoss, and the number of sampling in University-1652.D$\rightarrow$S means the task of Drone$\rightarrow$Satellite, and S$\rightarrow$D means the task of Satellite$\rightarrow$Drone.}
\label{table7}
\resizebox{1.0\hsize}{!}{
\begin{tabular}{c c c|c c}
\specialrule{0.75pt}{0pt}{0pt}   

\multirow{2}{*}{KLLoss} & \multirow{2}{*}{\shortstack{TripletLoss\\(M=0.3)}} & \multirow{2}{*}{\shortstack{Sampling\\Rate}} & \multicolumn{2}{c}{AP (\%)} 
\\
& & &{$\ \ \ \ $D$\rightarrow$S$\ \ \ \ $}&{$\ \ \ \ $S$\rightarrow$D$\ \ \ \ $}
\\
\specialrule{0.5pt}{0pt}{0pt}   
 {} & {} & {1$\times$} & {83.30} & {79.87} \\
 {\checkmark} & {} & {1$\times$} & {84.14} & {80.93} \\
 {\checkmark} & {\checkmark} & {1$\times$} & {84.82} & {81.53} \\
 {} & {\checkmark} & {1$\times$} & {84.85} & {81.52} \\
 {} & {\checkmark} & {2$\times$} & {86.36} & {82.69} \\
 {} & {\checkmark} & {3$\times$} & {\textbf{86.71}} & {\textbf{83.37}} \\
\specialrule{0.75pt}{0pt}{0pt}  
\end{tabular}
}
\end{table}

\textbf{Effect of some other tricks.} For the task of matching the drone-view and the satellite-view images, we adopt three tricks of KLLoss, TripletLoss with margin=0.3, and multiple sampling for the FSRA to improve the performance. As shown in Table VII. Only using KLLoss increases by 0.84\%/1.06\% AP on the task of Drone$\rightarrow$Satellite / Satellite$\rightarrow$Drone. Only using TripletLoss increases by 1.52\%/1.66\% AP. When we use KLLoss and TripletLoss at the same time, the accuracy of AP is not improved much. Thus, we did not use KLLoss but TripletLoss in our model. We guess that TripletLoss and KLLoss are consistent in the same direction of network fitting. In addition, we deploy the sampling strategy as a trick. Based on TripletLoss with margin=0.3. When the number of sampling reaches 2$\times{}$, the AP of FSRA increases by 1.51\%/1.17\% on the task of Drone$\rightarrow$Satellite / Satellite$\rightarrow$Drone. When the number of sampling reaches 3$\times{}$, the AP increases by 1.86\%/1.85\%. The performance improvement obtained by multiple sampling is due to the expansion of the data which can strengthen the fitting of the network and balance the resources from different domains.

\begin{figure}[!t]
\centering
\includegraphics[width=0.5\textwidth]{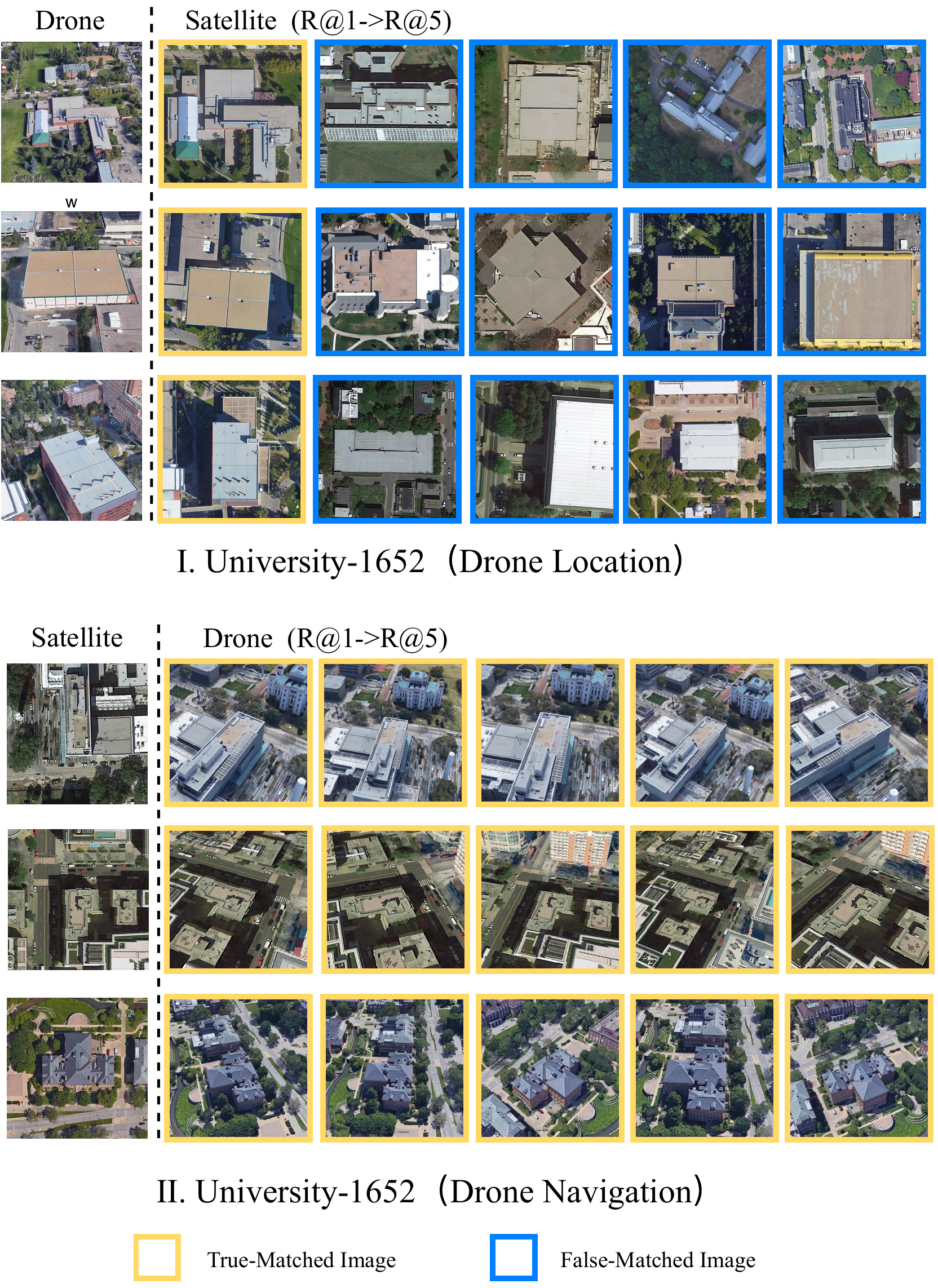}
\caption{Qualitative image retrieval results. (I) Top-5 retrieval results of drone view target localization on University-1652. (II) Top-5 retrieval results of drone navigation on University-1652. The yellow box indicates the true-matched image, and the blue box indicates the false-matched image.}
\label{fig_12}
\end{figure}

\subsection{Visualization Of Qualitative Result}
For the two basic tasks of the University-1652 dataset: drone view target localization and Drone Navigation, we visualize some retrieved results in Fig. \ref{fig_12}. We observe that FSRA can adapt to retrieving the available images from the gallery set in both drone view target localization and drone navigation tasks. In the task of drone view target localization, we randomly take out three drone-view images from the test dataset. For each drone-view image, we take out the top five similar images from the gallery set, and the FSRA obtains completely correct results as in Fig. \ref{fig_12} (I). In the Drone Navigation task, we randomly take out three Satellite-view images from the test dataset. For each Satellite-view image, we also take out the top 5 similar images in the gallery, because there is only one satellite image for each category. The proposed FSRA still achieved completely correct results as in Fig. \ref{fig_12} (II).

\section{Conclsion}
In this paper, we apply the structure of the Transformer to the field of cross-view geo-localization. The context information contained in the attention mechanism can distinguish more fine-grained features, and explore some associated information. Our experiments prove that the transformer-based FSRA can obtain state-of-the-art performance in the benchmark of the University-1652. In addition, some modules are proposed to improve model performance. HSM is proposed to implement patch-level semantic segmentation, and HAB is proposed to achieve region-level feature alignment. Although experiments shows that the proposed FSRA has strong robustness to feature misalignment and position shifts, there are still many parts that can be further improved. e.g., the structure of Vit can be modified to achieve more amazing performance. the backbone based on Vit-S has an increase in inference time compared to Resnet-50, which will be considered a shortcoming of this method. Besides, we also adopted a multiple sampling strategy to fit the model to a better state. This strategy can achieve a stunning rise, but the disadvantage is that it increases the training time. Finally, some other tricks such as mutual learning and TripletLoss are applied to make the FSRA stronger.
In the field of current geo-localization based on the perspective of drones. It is very necessary to construct a dense geographic dataset that the model can learn more distinctive and fine-grained features to achieve precise positioning. In the future, we will propose a new intensive UAV cross-view geo-localization dataset to meet the requirements of practical applications.

\section*{Acknowledgments}
The research is supported by the National Natural Science Foundation of China (Project 51605462).

{\appendices

\begin{table}[h]
\renewcommand\arraystretch{1.5}
\caption{The models were trained on ground, drone and satellite views, and the accuracy of matching between ground and drone views was tested. Where G refers to ground-view and D refers to drone-view.}
\label{table8}
\resizebox{1.0\hsize}{!}{
\begin{tabular}{c|c|ccc}
\specialrule{0.75pt}{0pt}{0pt}   

Model & Direction & R@1 & R@Top1\% &AP\\
\specialrule{0.5pt}{0pt}{0pt}   
\multirow{2}{*}{University-1652} & {G$\rightarrow$D} & {0.85} & {20.36} & {0.71} \\
& {D$\rightarrow$G} & {0.99}& {15.07} & {1.11}\\
\multirow{2}{*}{LPN} & {G$\rightarrow$D} & {0.85} & {20.47} & {0.94} \\
& {D$\rightarrow$G} & {1.70}& {17.41} & {1.70}\\
\multirow{2}{*}{FSRA(ours)} & {G$\rightarrow$D} & {1.94} & {31.91} & {1.67} \\
& {D$\rightarrow$G} & {2.75}& {24.92} & {2.63}\\
\specialrule{0.75pt}{0pt}{0pt}
\end{tabular}
}
\end{table}

\section{Does it work for ground view?}
The drone-view can be used as an intermediate view between the satellite and ground view because there is a 90 degree deviation between the ground-view image and the satellite-view image, and the occlusion between objects, the drastic differences in viewpoints and even the temporal gap between scenes make it very challenging. We try to use the drone-view to match the ground-view and thus indirectly reduce the difficulty of the matching between ground and satellite. For fairness, we apply the same learning strategy to the different models (all using only classification loss). The experimental results are shown in Table VIII. The proposed FSRA has improved somewhat compared with University-1652 and LPN, but still remains in single digits. Matching single-view ground images with UAV images is a huge challenge, mainly because there is a mismatch of shooting angles between ground-view images and UAV-view images, which in turn leads to large differences in the included content.

\begin{figure}[!t]
\centering
\includegraphics[width=0.5\textwidth]{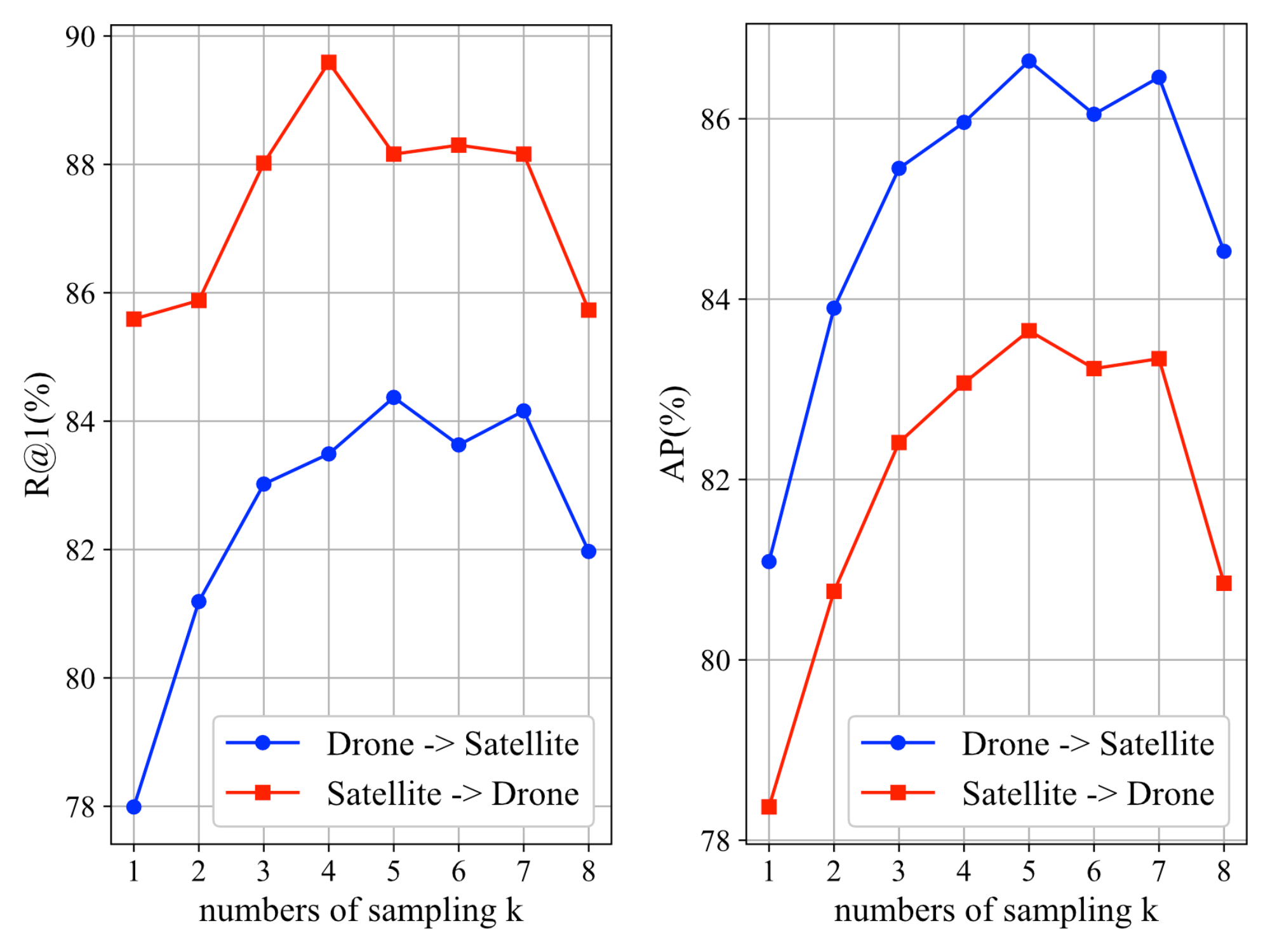}
\caption{The effect of the variation of numbers of sampling k on the final training results of the model when batchsize=16 is demonstrated, where the results of the R@1 evaluation metric are shown on the left and the results of the AP evaluation metric are shown on the right.}
\label{fig_13}
\end{figure}

\section{Does batchsize have effects on the choice of k?}
In order to verify the effect of batchsize on the choice of the optimal $k$ value, we increased the batchsize from 8 to 16 and conducted experiments for $k$ varying from 1 to 8 (to avoid the effect of TripletLoss positive and negative sample ratios on the experiments, only classification loss was used in the experiments). as shown in Fig. \ref{fig_13}, when the batchsize increases to 16, the optimal hyperparameter $k$ should be chosen to be around 5 ($k=3$ reaches the optimum for batchsize=8). This also verifies our statement in \emph{The impact of sampling on accuracy} that the proportion of samples of the same class in a batch affects the effectiveness of model training. Our proposed multiple sampling strategy can be used not only to expand the severely underrepresented satellite images in University-1652, but also to change the distribution of samples in the batch, so our proposed multiple sampling strategy needs to select the best $k$ value according to the actual batchsize. We conclude that $k=3$ is optimal when batchsize=8 and $k=5$ is optimal when batchsize=16. It should also be noted that increasing the value of $k$ increases the model training time exponentially, but does not have any effect on the inference process.
}

\bibliographystyle{IEEEtran}
\bibliography{IEEEabrv,FSRA}

\end{document}